\documentclass[a4paper,amsmath,amssymb]{jpconf}
\usepackage{graphicx}
\usepackage{hd3head}
\usepackage{amsmath, amsthm, amssymb}
\begin{document}
\title{Combining Machine Learning and Physics to Understand Glassy Systems}

\author{Samuel S. Schoenholz}

\address{Google Brain}

\ead{schsam@google.com}

\begin{abstract}
Our understanding of supercooled liquids and glasses has lagged significantly behind that of simple liquids and crystalline solids. This is in part due to the many possibly relevant degrees of freedom that are present due to the disorder inherent to these systems and in part to non-equilibrium effects which are difficult to treat in the standard context of statistical physics. Together these issues have resulted in a field whose theories are under-constrained by experiment and where fundamental questions are still unresolved. Mean field results have been successful in infinite dimensions but it is unclear to what extent they apply to realistic systems and assume uniform local structure. At odds with this are theories premised on the existence of structural defects. However, until recently it has been impossible to find structural signatures that are predictive of dynamics. Here we summarize and recast the results from several recent papers offering a data driven approach to building a phenomenological theory of disordered materials by combining machine learning with physical intuition.
\end{abstract}

\section{Introduction}
\label{sec:Introduction}
Over the last century, enormous progress has been made towards understanding physical systems in their condensed phases. For simple liquids, theoretical developments have resulted in liquid state theory~\cite{henderson69} that assumes the system to have uniform local structure. For crystalline solids, theories based on the existence of topological defects that couple elastically through the bulk have been remarkably successful~\cite{merminbook,landau1986}. Across both fields rapidly advancing numerical simulations of materials have typically been used to corroborate theories rather than inform novel theoretical work. 

By contrast, a comprehensive theory of glasses and supercooled liquids has eluded researchers. In part, the struggle to understand these systems stems from two factors. First, amorphous systems have a large number of degrees of freedom and there are few heuristics that one can use to determine which (if any) may be neglected. Second, the range of temperatures where glassy liquids are in equilibrium is small since viscosity increases so rapidly with decreasing temperature. It is therefore difficult to test theories without grappling with non-equilibrium effects. Together these two effects have resulted in a field with too many theories that are not sufficiently constrained by experiments. 

Mode coupling theory~\cite{Reichman2005} and random first order theory~\cite{kirkpatrick1987,parisi10,berthier11,charbonneau14} extend liquid state theory to formally describe super-cooled liquids and glasses at the level of mean field theory. However, it is unclear to what extent these mean field results survive down to finite-dimensional realistic glass formers. Indeed, the dynamical glass transition predicted by mode-coupling theory is conspicuously absent in three-dimensions~\cite{gotze1999}. By contrast, theoretical frameworks such as shear transformation zones~\cite{falk98} and dynamical facilitation~\cite{fredrickson44} are premised on the existence of defects analogous to those in a crystal. Placing these theories on firm experimental footing has been difficult because historically, only weak correlations were found between measures of local structure - such as bond orientational order~\cite{kawasaki14}, free volume~\cite{manning11}, or local potential energy~\cite{royall08} - and dynamics.

In this article we will review a series of papers~\cite{cubukschoenholz15, schoenholzcubuk16, cubukschoenholz16, schoenholzcubuk17} that combine traditional physical intuition with machine learning to analyze numerical simulations of molecular glasses and supercooled liquids and gain novel physical insight. To do this we will first introduce a quantity that we term the ``softness''. The softness leverages large amounts of data from molecular dynamics simulations or experiment to design a function of local-structure that most strongly correlates with dynamics. We proceed to investigate the relationship between softness and dynamics. Finally we leverage this connection to approach several open problems in glassy physics.
 
\section{Background}

We begin by presenting a brief recapitulation of the phenomenology of the glass transition. For a more detailed account see i.e. Debeneditti and Stillinger~\cite{stillinger01}. The freezing of a liquid into a crystalline solid proceeds via a first-order phase transition. At a critical temperature, $T_m$, the free energy of the disordered, isotropic, and homogeneous arrangement of particles in the fluid becomes equal to the free energy of the ordered anisotropy of the crystal. The ordered phase proceeds to nucleate in the fluid until the entire system freezes. In the view of Landau this process is relatively well understood and is described by the breaking of the continuous rotational and translational symmetries of the fluid down to the discrete set of crystallographic symmetries. 

However, the process of nucleation is relatively slow. If the liquid is cooled quickly enough, crystallization can be avoided and the system enters a long-lived metastable state known as the supercooled liquid. As the temperature of the supercooled liquid is lowered further it becomes increasingly sluggish. At some temperature, known as the glass transition temperature $T_g$, the time required for the system to equilibrate exceeds any reasonable experimental timescale and it is said that the liquid has solidified to a glass. However, unlike in the case of the liquid to crystal transition there are no diverging length scales, critical exponents, or structural changes that have been associated with this change of state. Moreover, since the system has fallen out of equilibrium, the usual tools of statistical mechanics may not be applied once the temperature is lower than $T_g$. Finally, it is well known that the specifics of the glass transition depend on the cooling protocol: the slower the cooling rate the longer the system is able to remain equilibrated and the lower the glass transition temperature. 

One of the biggest mysteries associated with these systems is the extremely rapid increase of the relaxation time (or viscosity) as temperature is lowered. Many systems feature a viscosity that increases by nearly ten orders of magnitude as $T_g/T$ varies by only about $10\%$ and which is accompanied by no obvious structural change. Glass formers are generically categorized into two groups based on the manner in which their viscosity varies with $T_g/T$. In strong glass-formers, such as SiO$_2$, viscosities scale exponentially with $T_g/T$ as $\eta\sim\exp(A/T)$ and $A$ can be interpreted as an energy scale. However, the many so-called ``fragile'' glass formers slow down significantly more quickly.  The form of the viscosity in fragile systems is unknown and moreover it is not known whether or not the viscosity diverges at a genuine phase transition. As discussed above, this leads to a situation where the available data is insufficient to constrain the space of possible models.

A second phenomenon that supercooled liquids and glasses exhibit is non-exponential relaxation. To quantify the manner in which glasses relax one can use either the self-intermediate scattering function, $F_s(q,t)$, or the overlap function defined by the equation,
\begin{equation}\label{eq:overlap}
q(t) = \frac1N\sum_i\Theta(|r_i(t)-r_i(0)| - a).
\end{equation}
The overlap function measures the fraction of the system that has moved greater than a characteristic distance, $a$ after a time $t$. Both the overlap function and the self-intermediate scattering function behave similarly. In simple liquids, particles exhibit Brownian motion and $q(t)$ decays exponentially with a characteristic timescale known as the $\alpha$-relaxation time. However, as supercooled liquids get closer to their glassy state the decay of the overlap function becomes stretched as $q(t)\sim e^{-(t/\tau_\alpha)^\beta}$. The origin of this stretched exponential form still lacks solid explanation in three-dimensions.

A final open question concerns the increasingly heterogeneous dynamics that emerge as supercooled liquids are cooled towards their glassy state (for a more thorough discussion see i.e. Keys \textit{et al.}~\cite{keys13} or Widmer-Cooper \textit{et al.}~\cite{cooper09}). In a simple liquid the activity of different regions of the system are statistically identical and dynamics are homogeneous. By contrast in the supercooled liquid and glassy phase there are large heterogeneities in the mobility of particles. The origin of these large differences in particle motion is unknown and, indeed, it has yet to be determined whether or not they are a result of structural defects or whether they are purely dynamical in nature.

A number of other approaches have attempted to identify a structural signature of dynamics in supercooled liquids. These methods traditionally are divided into two approaches. The first approach correlates quasilocalized vibrational modes in the dynamical matrix with rearrangements~\cite{cooper09, manning11, schoenholz14, rottler14}. However, this method has a number of problems. First, diagonalizing the dynamical matrix is costly and is only applicable to systems where the Hamiltonian is available. Second, coupling between localized regions and the elastic background obfuscates the identification of defects. A second approach attempts to define locally preferred structures by hand and identify them in the glassy liquid~\cite{kawasaki14,gilman75,berthier07,royall08,jack14}. However, the identification of locally preferred structures is far from trivial and must be done by hand for each new system. Both methods typically feature correlations between structure and dynamics that are too low to use in the construction of a theory. Several more recent papers have attempted to approach this question using the notion of local thermal energy~\cite{Zylberg17}.

\section{Softness}

We will summarize the results from four papers~\cite{cubukschoenholz15, schoenholzcubuk16, cubukschoenholz16, schoenholzcubuk17} that discuss the three questions posed in the previous section by exploring the relationship between local-structure and dynamics in supercooled liquids and glasses. At the same time, we place some emphasis on being pedagogical. Our primary tool in this investigation will be a quantity that we introduced in two papers~\cite{cubukschoenholz15, schoenholzcubuk16} and term the ``softness''. We will describe the construction of the softness presently.

Suppose we have access to the time-dependent positions of particles in a glassy liquid, $\{x_i(t)\}$, either from computer simulation or experiment. Qualitatively, it seems that at sufficiently low temperatures, the position of particles in a supercooled liquid spend their time fluctuating in the vicinity of a local-minimum in the energy landscape. Thermal fluctuations intermittently drive the system over saddle-points to neighboring minima. It has been observed~\cite{keys11} that these transition events correspond to localized rearrangements involving a small fraction of the system. These intermittent dynamics can be seen at the single particle level in fig.~\ref{fig:phop} (a). If there is a connection between local-structure and dynamics then the local-neighborhood of particles just before a rearrangement should be distinguishable the that of particles that have not rearranged in a significant amount of time.  

Instead of trying to intuit the relationship between structure and dynamics, we will take a machine learning approach using the large amounts of data from either molecular dynamics or experimental data~\cite{chen13}. We will take the following steps:
\begin{itemize}
\item Identify a population of local regions that are about to experience rearrangements and another population of regions that are unusually stable.
\item Parametrize the degrees of freedom of each local region in a manner that is amenable to analysis.
\item Learn the function of the parametrization that optimally separates the rearranging population from the stable population. 
\end{itemize}
Each step outlined here can be performed in a number of ways. We find that our results are remarkably insensitive to the specific choices made. In light of this, we will choose to present a construction of the softness field that is easy to implement and physically motivated. This incurs a slight loss in our ability to differentiate between the two groups ($\sim2\%$ difference in classification accuracy). For more details see the supplementary material in Schoenholz and Cubuk \textit{et al.}~\cite{schoenholzcubuk16}.

Throughout this paper we will focus on results from simulations of a 10,000-particle, 80:20 binary Lennard-Jones (LJ) Kob-Andersen mixture~\cite{kob94}. We measure time in units of $\tau = \sqrt{\epsilon_{AA}/\sigma_{AA}^2}$ and take Boltzmann's constant to be $k_B =1$. Finally, we cut off the LJ potential at $2.5\sigma_{AA}$ and smooth it so that the force varies continuously. Simulations were performed in LAMMPS~\cite{LAMMPS} using and NVT ensemble with a Nos\'e-Hoover chain thermostat and a timestep of $0.0025\tau$. Every $\tau$-timesteps we quench the system to its nearest inherent structure using a combination of conjugate gradient and FIRE minimization. We use inherent structure positions instead of instantaneous positions. However, as with the details of our approach our results are largely insensitive to these choices and indeed time-averaged positions or even instantaneous positions may be used in its place~\cite{cubukschoenholz15}. While we will not discuss it in detail here our method has also proven successful when applied to experimental systems and polymeric glasses~\cite{sussman16}.

\subsection{Identifying Rearrangements}

To identify rearrangements we have used both $D^2_{\text{min}}$ as introduced by Falk and Langer~\cite{falk98} as well as $p_{\text{hop}}$ discussed in Candelier \textit{et al.}~\cite{candelier10, smessaert13}. While $D^2_{\text{min}}$ is better suited for systems that are under external load, $p_{\text{hop}}$ gives slightly better results for quiescent systems and so we will focus on it here.

We define $p_{\text{hop}}$ by first picking a timescale $t_R=10\tau$ that is approximately the time it takes for the system to complete a rearrangement. At a time $t$, this choice then gives two intervals $A = [t-t_R/2, t]$ and $B=[t,t+t_R/2]$. With reference to these two intervals we define for a single particle,
\begin{equation}\label{eq:phop}
p_{\text{hop}}(i,t) = \sqrt{\langle (x_i - \langle x_i\rangle_B)^2\rangle_A\langle (x_i - \langle x_i\rangle_A)^2\rangle_B}.
\end{equation} 
where $\langle\rangle_A$ and $\langle\rangle_B$ are averages over the intervals A and B respectively. In practice these averages are computed as $\langle f(x_i)\rangle_A = \frac2{t_R}\sum_{t'=t-t_R/2}^t f(x_i(t'))$ with an analogous expression holding for $\langle f(x_i)\rangle_B$.

\begin{figure}[!h]
\centering
\includegraphics[width=0.9\linewidth]{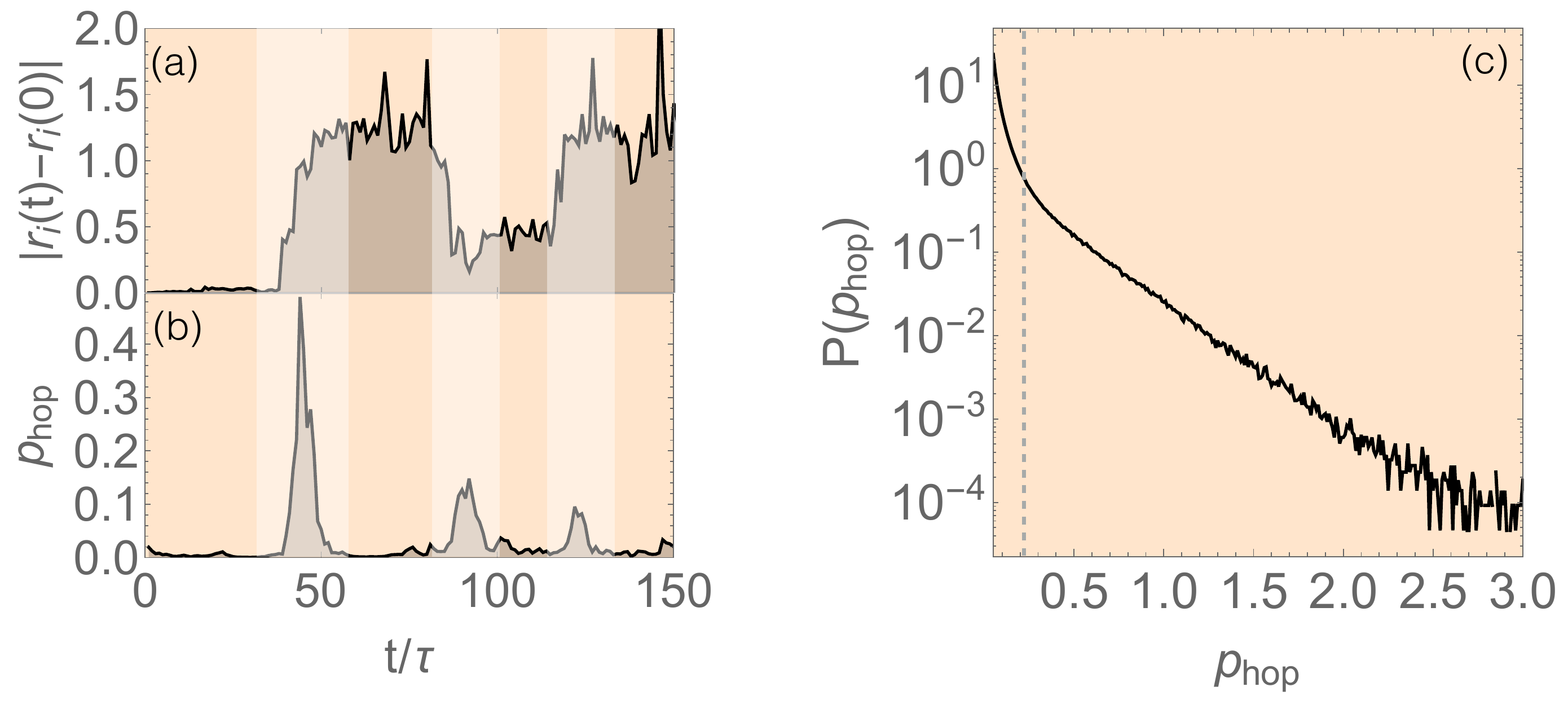}
\caption{Quantifying rearrangements. (a) The distance in the inherent structure position of a particle as a function of time. Several well-defined hopping events can be seen. (b) The $p_{\text{hop}}$ indicator function of this same trajectory. We see that $p_{\text{hop}}$ is very sensitive to rearrangements of the inherent structure position. (c) The probability distribution of $p_{\text{hop}}$. We see a clear crossover to an exponential distribution at a well-defined value of $p_{\text{hop}}$.}
\label{fig:phop}
\end{figure}

If no rearrangement takes place between intervals $A$ and $B$ then eq.~\eqref{eq:phop} reduces to computing the variance of the particle position over time, which will be intimately related to the volume of caged motion that the particle experiences. In the event that the inherent structure is used instead of the instantaneous position this amounts to computing the variance of the inherent structure position (which will be nonzero both because the the inherent structure position might not be exactly constant and because the minimization has some inherent uncertainty associated with it). If a rearrangement does occur then $p_{\text{hop}}$ will be proportional to the square of the distance the particle moves as the state transitions between the two minima. The behavior of $p_{\text{hop}}$ over the course of a rearrangement can be seen in fig.~\ref{fig:phop} (a) and (b).

Together these two limits result in a separation of scales that emerge between rearrangements and caged motion. This can be seen by referring to the distribution of $p_{\text{hop}}$ in fig.~\ref{fig:phop} (c) which exhibits a clear crossover between exponential and non-exponential behavior. This allows us to cleanly identify rearrangements in disordered systems.

To do this we select two thresholds $p_{L}\approx 0.05$ which defines a lower bound on what we consider to be a rearrangement. We then identify, from one or a series of particle trajectories, all rearrangement events which we define to be instances where $p_{\text{hop}}$ exceeds $p_L$. For each event we define $p^*$ to be the maximum value of $p_{\text{hop}}$ achieved during the event. We also define $t_{\text{start}}$ and $t_{\text{end}}$ to be the time when $p_{\text{hop}}$ first is greater and less than $p_{L}$ respectively. Empirically this gives a very precise definition of rearrangements. One can then study quantities such as the duration of rearrangements by considering $t_{\text{end}} - t_{\text{start}}$ or the displacement of particles during rearrangements, $\Delta x_i = x_i(t_{\text{end}}) - x_i(t_{\text{start}})$.

With rearrangements well-defined we construct two populations of particles. The first group are particles that are about to experience a large rearrangement. We define large rearrangements to be those with $p^* > p_c\approx 0.6$. For each event we consider the particle configuration just before the rearrangement at $t_{\text{start}} - 2\tau$. The second group contains particles that have not rearranged in a long time. Thus we take particles that experience no rearrangement events for approximately $\tau_c \approx \tau_\alpha$. Here we consider particle configurations in the middle of the interval. In general the larger $p_c$ and $\tau_c$ the better the results so long as each population of particles contain at least 3000 particles.

\subsection{Parametrizing Local Structure}
 
We now present a simple and physically motivated parametrization of the local-structure around a central atom $i$. There are, however, a wide-variety of methods that can be used instead~\cite{behler07, cubuk2014theory, cubuk17, gilmer17, faber17}. Qualitatively, our results are insensitive to the choice of parametrization so long as it is sufficiently expressive. We typically find that using more sophisticated methods for describing local-structure results in an improvement in our ability to differentiate the two populations by about $2$-$4\%$. 

\begin{figure}[!h]
\centering
\includegraphics[width=0.9\linewidth]{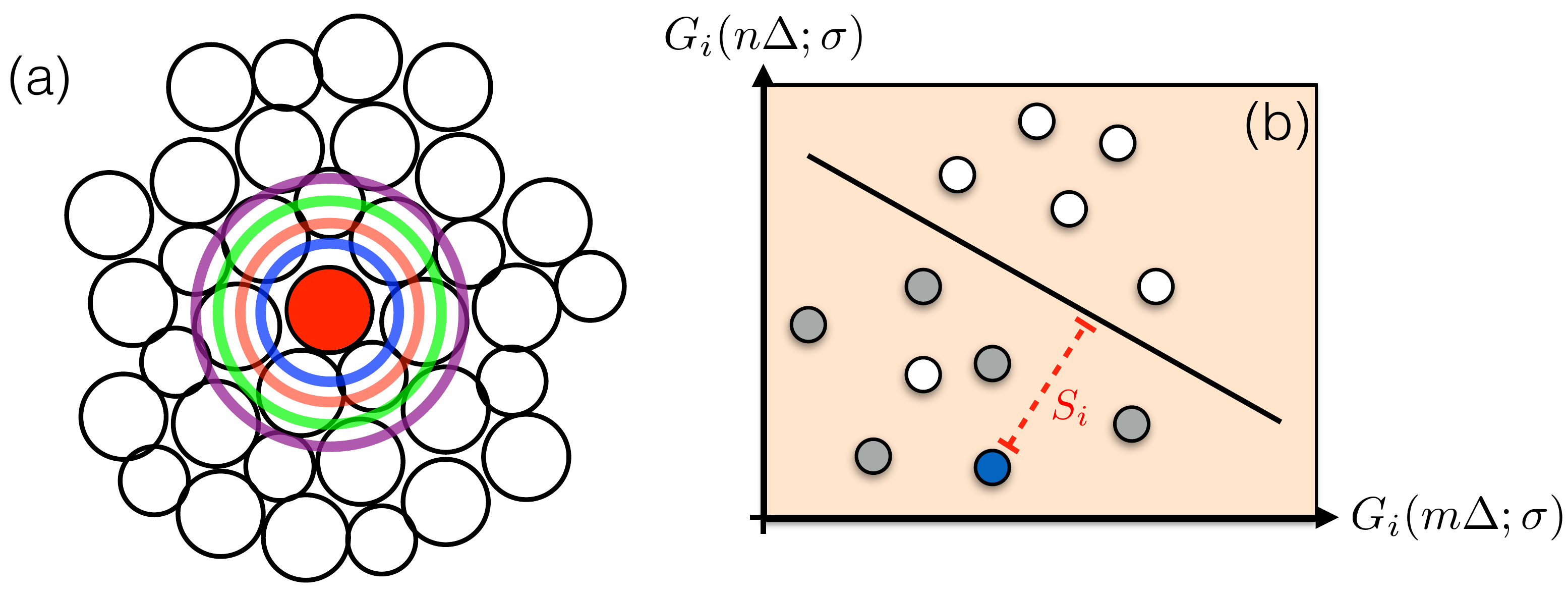}
\caption{Parametrizing local structure. (a) We show schematically the functions $G^X_i(r;\sigma)$ for several different values of $r$. (b) A schematic showing how the local-structure around a central atom is embedded into a high dimensional space. We then find a hyperplane that separates particles that are about to rearrange from those that have not rearranged in a long time. Once the hyperplane has been established we call the softness the distance of that local configuration to the hyperplane.}
\label{fig:symmetry}
\end{figure}

We begin by defining the function,
\begin{equation}
G_i^X(r;\sigma) = \frac 1{\sqrt{2\pi}}\sum_{j\in X}e^{-\frac1{2\sigma^2}(R_{ij} - r)^2}
\end{equation}
where $R_{ij} = |x_i - x_j|$ is the distance between particles $i$ and $j$ and $X\in\{A,B\}$ refers to the species of atom. $G_i^X(r;\sigma)$ essentially counts the number of particles of a given species whose distance from a central particle is within about $\sigma$ of $r$. This can be seen schematically in fig.~\ref{fig:symmetry} (a). In principle if we could access the entire function $G_i(r;\sigma)$ we would be able to perform inferences on the local structure that depended on the entire density profile around an atom. 

In practice we discretize the set of radii into $r_n = n\Delta$ and consider the vector $G_i = (G^A_i(\Delta;\sigma), G^A_i(2\Delta;\sigma),\cdots, G_i^B(N\Delta;\sigma))\in\mathbb R^N$. This provides an embedding of the density profile into $\mathbb R^N$ where many more techniques have been developed to help perform inference. A good selection of $\Delta$ that we found is $\Delta\approx \sigma$ and $\sigma\approx 0.1\sigma_{AA}$. However, note that with this selection $G^X_i(n\Delta;\sigma)$ and $G^X_i((n+1)\Delta;\sigma)$ will be highly correlated. We will discuss the implications of this correlation more in the next section. By choosing $\Delta \approx 3\sigma$ we get a set of $G_i^X(n\Delta;\sigma)$ that are not as correlated with a minimal drop in accuracy (usually $\lesssim 1\%$). 

\subsection{Constructing the Softness}

With a population of rearranging and stable particles defined and a parametrization of the local-structure in the neighborhood of each particle through an embedding into $\mathbb R^N$, we can proceed to discuss the construction of the softness field itself. The question of finding functions of features (in this case $G_i$) that best separates points in different classes (in this case about to rearrange and stable) is well studied in statistics and machine learning. We will use a method from machine learning called support vector machines~\cite{SVM} with a linear kernel, however as with most other pieces of the method a number of different techniques can be used with similar effect (e.g. logistic regression, nonlinear kernels, neural networks).

Using linear support vector machines gives us a hyperplane defined by a normal $\hat w$ and a bias $b$ so that $\hat w\cdot G_i + b$ is positive when a particle is about to rearrange and negative when a particle is stable. For a schematic representation see fig.~\ref{fig:symmetry} (b). When we apply this technique to systems at a temperature $T=0.47$ we find that our hyperplane is able to correct separate about $88\%$ of the points into their correct classes.

\begin{figure}[!h]
\centering
\includegraphics[width=0.9\linewidth]{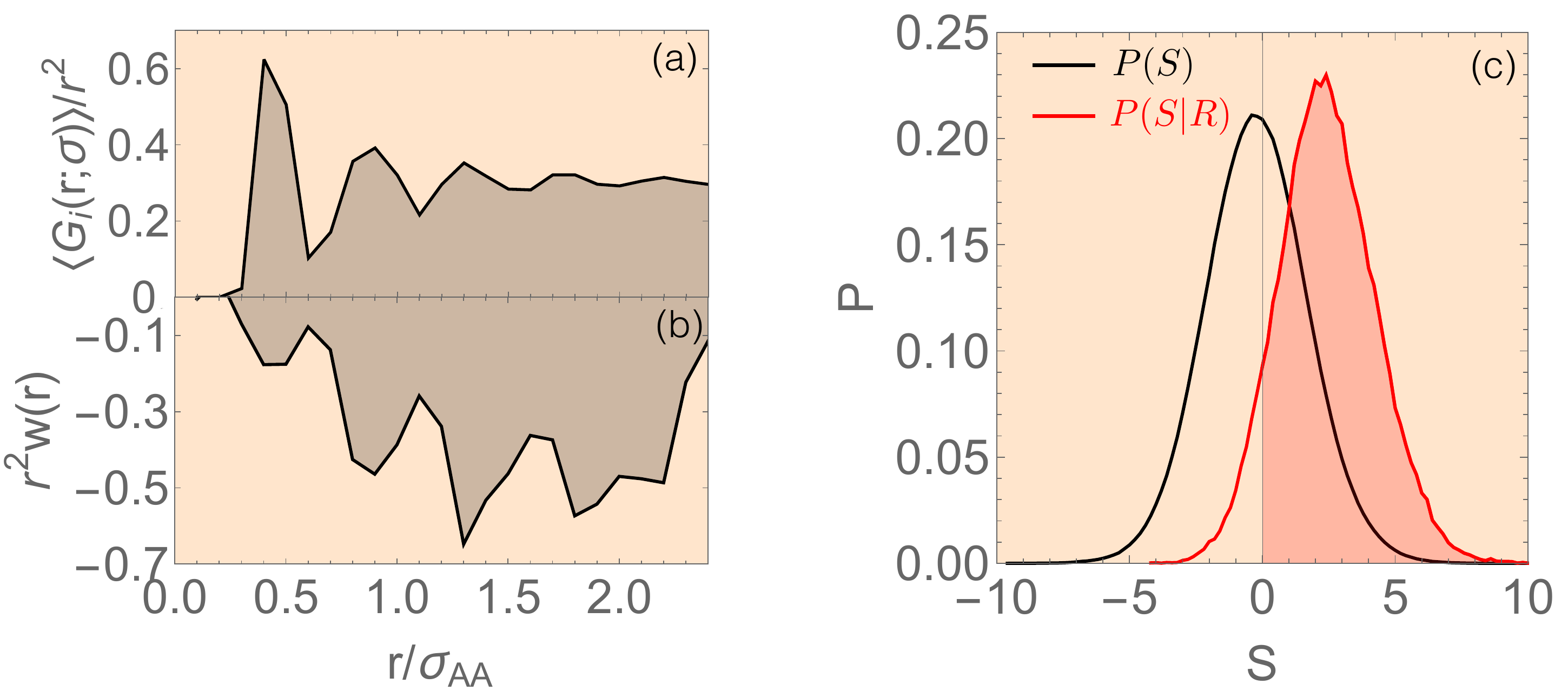}
\caption{The softness. (a) The function $r^{1-d}\langle G_i(r;\sigma)\rangle$ averaged over a whole system. We see that it is qualitatively similar to $g(r)$. (b) The function $\tilde w(r)$ learned by the support vector machine. We see that it is sensitive to $g(r)$ up to about $2.5\sigma_{AA}$. (c) The softness distribution and the softness distribution for particles that were about to rearrange. We see that the two distributions are well-separated and $90\%$ of the rearranging particles get assigned $S>0$.}
\label{fig:softness}
\end{figure}

Once this has been done new local structures can be classified using this hyperplane and we can make predictions about whether or not the particle is about to rearrange from its structure. However, we found it more useful to consider the signed-distance to the hyperplane, $S_i = \hat w\cdot G_i + b$ which we call the softness. The rest of this paper will be concerned with understanding properties of the softness.

One motivation for our simplified choice of parametrization as well as our choice of linear support vector machine as opposed to some more complicated nonlinear method is interpretability. We will now provide a physical picture for softness that is related, on average, to the radial distribution function. For the purposes of this discussion we will let $\Delta = \sigma$ and we will consider a hypothetical system composed of only one species of atom however the result generalizes trivially to multiple species. Let us also assume we fix the maximum radius at which we probe local structure so that the number of features is $N = R_c/\sigma$ with $R_c$ the radial cutoff. It follows that, neglecting the bias, the softness can be written as,
\begin{equation}S_i = \sum_{n=1}^{N} G_i(n\sigma;\sigma)w(n\sigma)\end{equation}
where $w(i)$ are the weights from the SVM. In the limit of zero width we find that,
\begin{align}
S_i &= \lim_{\sigma\to0}\sum_{n=1}^{R_c/\sigma} G(n\sigma;i,\sigma)w(n\sigma)\\
&=\lim_{\sigma\to0}\sum_{n=1}^{R_c/\sigma}\sigma\left[\frac1{\sqrt{2\pi\sigma^2}} \sum_je^{-(R_{ij}-n\sigma)^2/2\sigma^2}\right]w(n\sigma).
\end{align}
This is exactly the definition of a Riemann integral and so we may write,
\begin{equation}
S_i =\int dr \left[\sum_{j}\delta(r-R_{ij})\right]w(r)
\end{equation}
where we have made the assignments $\lim_{\sigma\to0}\sum_{n=1}^{R_c/\sigma}\sigma\to\int_0^{R_c} dr$, $n\sigma\to r$, and $\lim_{\sigma\to0}\frac1{\sqrt{2\pi\sigma^2}} e^{-x^2/2\sigma^2} \to \sqrt{2\pi}\delta(x).$

We can use this formula to understand the softness in two different but equivalent ways. We define the ``local density'' operator ,
\begin{equation}
g_i(r) = \frac1{r^{d-1}}\sum_{j}\delta(r-R_{ij})
\end{equation}
that measures the density of particles a distance $r$ away from a tagged particle $i$. In this case, the softness can be written as,
\begin{equation}
S_i = \int dr g_i(r)\tilde w(r)
\end{equation}
where we have defined $\tilde w(r) = r^{d-1}w(r)$. It follows that we can think of $S_i$ as the functional that integrates the local density weighted by some weight function $\tilde w(r)$. The purpose of the SVM is to then find an appropriate weight function over the local density. We note further that - when the system is isotropic - our symmetry functions are related to the radial distribution function in the sense that $g(r) \sim \lim_{\sigma\to0}r^{1-d}\langle G_i(r;\sigma)\rangle$. 

These two relationships can be seen in fig.~\ref{fig:softness} (a) and (b). In (a) we see that when $\sigma$ is sufficiently small $r^{1-d}\langle G_i(r;\sigma)\rangle$ does indeed give the radial distribution function. In (b) we see the weight function, $r^{d-1}w(r)$, that is learned by our network. We see that the weight function features many of the characteristic features of $g(r)$. Moreover, we see that $r^{d-1}w(r)$ has support only within a region where $r\lesssim2.5\sigma_{AA}$, this gives a characteristic volume that is important for predicting rearrangements. 

To investigate the effectiveness of softness at predicting rearrangements it is useful to consider the distribution of softnesses over an entire molecular dynamics simulation, $P(S)$, and to compare it with the distribution of softness for those particles that are currently involved in a rearrangement, $P(S|R)$. This is shown in fig.~\ref{fig:softness} (c). We see, in particular, that both $P(S)$ and $P(S|R)$ are approximately Gaussian distributed. Moreover we notice that the classification accuracy of our model is given by $P(S>0|R)\approx 0.88$ which means that our model is able to correctly predict approximately $90\%$ of rearrangements correctly.

\section{Relating Softness with Dynamics}

\subsection{The Probability of Rearrangement}

The success of support vector machines at differentiating particles that are about to rearrange from those that are stable offers definitive proof that there is a strong structural component to dynamics in glassy systems. However, the definition of softness unto itself is not terribly informative. As with previous predictors of dynamics from local structure it is not obvious how to relate classification accuracy with a useful theory of dynamics of glassy systems. This issue is further complicated by the fact that softness itself is not a physical quantity. Indeed, both the scale and the shift of the hyperplane identified by the support vector machine are arbitrary and we could easily define a transformation of softness $\tilde S_i = \alpha S_i + \beta$ that would have essentially the same physical significance. 

\begin{figure}[!h]
\centering
\includegraphics[width=0.9\linewidth]{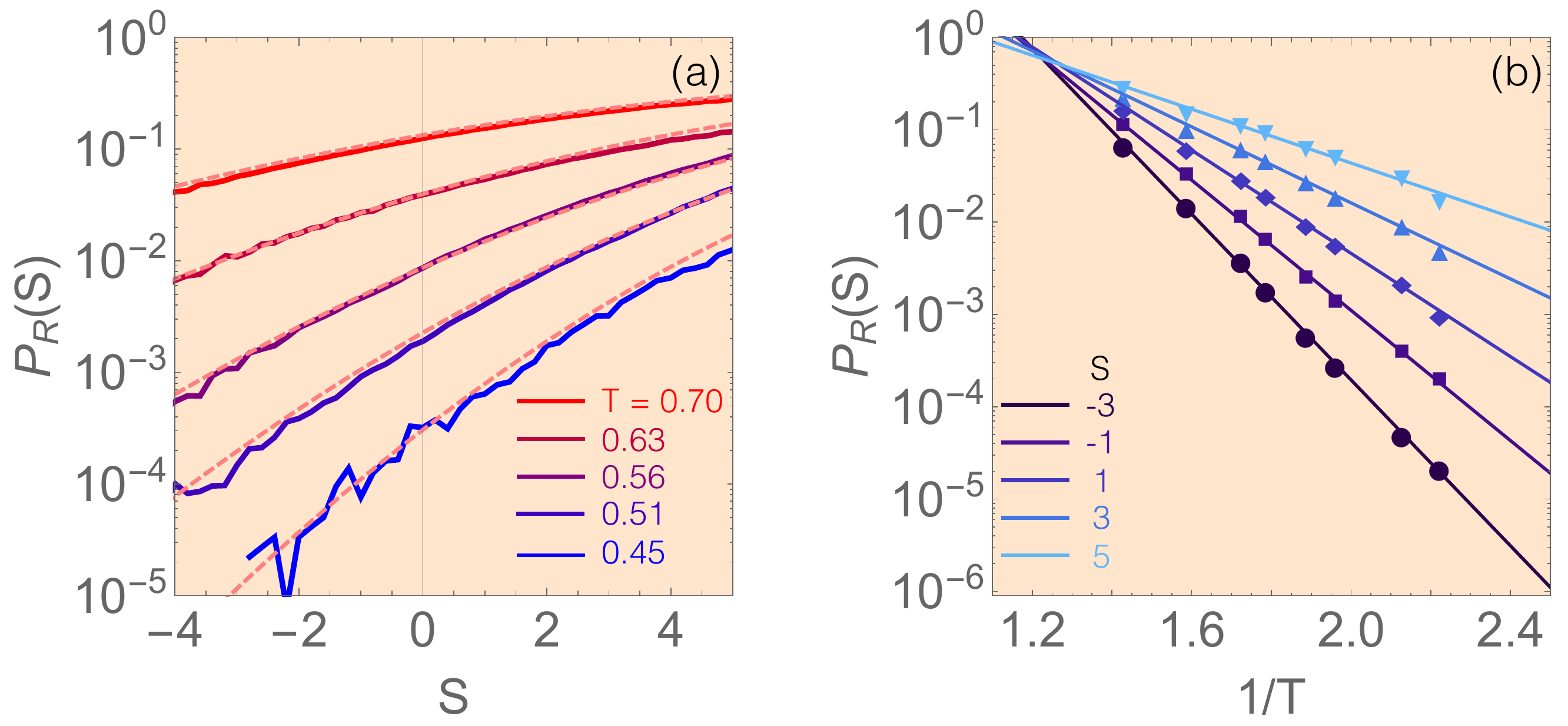}
\caption{The probability of rearrangement. (a) We plot the probability of rearrangement at different temperatures. Overlaid is the prediction from the Arrhenius form of $P_R(S)$. (b) Plotting $P_R(S)$ for different softness values as a function of $1/T$. We see that the probability of rearrangement viewed in this way is approximately Arrhenius.}
\label{fig:rearrangement_probability}
\end{figure}

There is a far more useful quantity that we can construct from an understanding of $P(S)$ and $P(S|R)$. Indeed using Bayes theorem we can compute the probability that particles of a given softness will rearrange,
\begin{equation}
P_R(S) \equiv P(R|S) = \frac{P(S|R)P(S)}{P(R)}.
\end{equation}
This allows us to predict, from local-structure, the probability that a particle will experience a rearrangement. In practice it is usually easier to compute $P_R(S)$ by evaluating using $p_{\text{hop}}$ at each instant of the simulation what fraction of particles are rearranging binned by softness. To define rearrangements here we can be less restrictive than when constructing the training set and take $p_{\text{hop}} \gtrsim 0.2$. We will briefly discuss the choice of cutoff later (and it is discussed more thoroughly in the supplementary material of Schoenholz \textit{et al.}~\cite{schoenholzcubuk16}) but the results are qualitatively insensitive to the selection. Note that $P_R(S_i)$ will be invariant to the transformation discussed above. 

We plot the probability of rearrangement at a number of different temperatures in fig.~\ref{fig:rearrangement_probability} (a). We see that there is a strong softness dependence on the probability of rearrangement with high-softness particles having a probability of rearrangement nearly two orders of magnitude larger than their low-softness counterparts. 

The probability of rearrangement can be interpreted in a slightly different way by plotting $P_R(S)$ against $1/T$ for different values of softness. The result of this transformation can be seen in fig.~\ref{fig:rearrangement_probability} (b) and we see, in particular, that for a given value of softness $P_R(S)$ depends exponentially on $1/T$. Surprisingly, this implies that, separated by softness, the probability of rearrangement is Arrhenius. With reference to transition state theory we can therefore write down what we believe to be a fundamental equation of glassy dynamics,
\begin{equation}\label{eq:rearrangement_probability}
P_R(S) = Z(S)e^{-E(S)/kT}.
\end{equation}
Here $E(S)$ and $Z(S)$ is a structure-dependent energy scale and multiplicity scale respectively that together govern local rearrangements. We will frequently refer to a structure dependent entropy, $\Sigma(S) = \log Z(S)$. 

\begin{figure}[!h]
\centering
\includegraphics[width=0.9\linewidth]{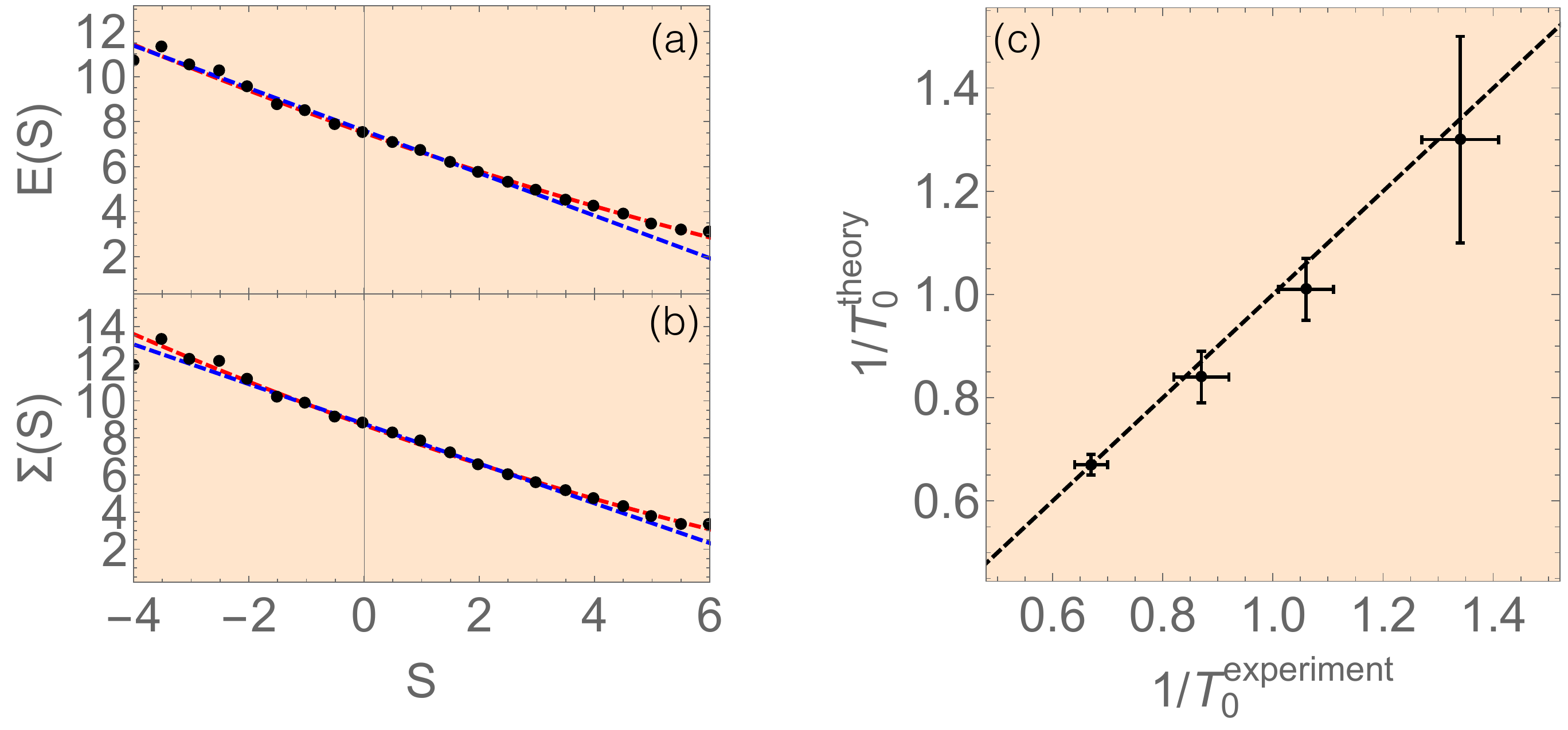}
\caption{Energy scales. (a) The energy scale governing local rearrangements as a function of softness with a linear fit (blue) and a quadratic fit (red). (b) The entropy scale $\Sigma$ also with linear and quadratic fits. (c) A comparison between the theoretical onset temperature and the onset temperature identified by Keys \textit{et al.}~\cite{keys11}. }
\label{fig:onset}
\end{figure}

Using eq.~\eqref{eq:rearrangement_probability} we can extract energy and entropy scales that control local rearrangements as a function of softness by fitting to the Arrhenius form in fig.~\ref{fig:rearrangement_probability} (b). The result of these fits can be seen in fig.~\ref{fig:onset} (a) along with linear (blue) and quadratic (red) approximations to $E(S)$ and $\Sigma(S)$. We see that both the energy scale and entropy scale are approximately linear near $S=0$. This is corroborated by considering the fits themselves; we find, $E(S)\approx E_0 - E_1 S + E_2 S^2$ with $E_0 \approx 7.50$, $E_1\approx  0.90$, $E_2\approx 0.02$ giving $E_1/E_2\sim 50$. This implies that nonlinear effects are relatively small in the regime with most of the probability mass of $S$. We will therefore typically take a linear approximation to the energy and entropy scales. We also find that $\Sigma(S) \approx \Sigma_1 - \Sigma_1 S + \Sigma_2 S^2$ with $\Sigma_0 \approx 8.70$, $\Sigma_1 \approx 1.1$, $\Sigma_2\approx 0.029$ which gives a similar ratio. In fig.~\ref{fig:rearrangement_probability} (a) we show $P_R(S)$ predicted by the quadratic approximation to the energy scale and entropy scale combined with the Arrhenius form in eq.~\eqref{eq:rearrangement_probability}.

For Lennard-Jones glasses we therefore have that,
\begin{equation}
P_R(S) \approx \exp\left[E_0\left(\frac1{T_0} - \frac1T\right)-E_1\left(\frac1{T_1}-\frac1T\right)S\right]
\end{equation}
where $T_0 = E_0/\Sigma_0$ and $T_1=E_1/\Sigma_1$. Note that $T_0 \approx T_1 \approx 0.86$ and so we can further approximate eq.~\eqref{eq:rearrangement_probability} by,
\begin{equation}\label{eq:rearrangement_probability_lj}
P_R(S) \approx \exp\left[\left(\frac1{T_0} - \frac1T\right)(E_0-E_1S)\right].
\end{equation}
When $T = T_0$ we find that $P_R(S) = 1$ independent of structure and the structural component of dynamics vanishes. We can compare this temperature with the onset of glassy dynamics found independently~\cite{keys11}. We see the result of this comparison in fig.~\ref{fig:onset} (b) for the Lennard-Jones mixture at densities, $\rho=1.15, 1.20, 1.25,$ and $1.30$. The excellent agreement between these two quantities lead us to the conclusion that we can actively define the onset of glassy dynamics by the onset of correlations between local-structure and dynamics. In addition to providing an explanation for the onset temperature, eq.~\eqref{eq:rearrangement_probability_lj} also explains the heterogeneous dynamics of glassy liquids as the heterogeneous distribution of softness leads a corresponding spatial distribution of local energy barriers to rearrangement.

\subsection{Relaxation}

We will now discuss the relationship between softness and relaxation in glassy systems both in- and-out- of equilibrium. To do this we first establish a connection between the probability of rearrangement and the relaxation time of a glassy liquid. We begin by defining a per-softness overlap function in analogy to eq.~\eqref{eq:overlap} to be the overlap after a time $t$ for particles that had softness $S$ at time $t=0$,
\begin{equation}\label{eq:softness_overlap}
q(S,t) = \frac1{N_S}\sum_i\Theta(|r_i(t) - r_i(0)| - a)\delta(S_i(0)-S)
\end{equation}
where $N_S$ is the number of particles with softness $S$ at $t=0$. The full overlap function is related to the softness dependent overlap function by $q(t) = \int dS q(S,t)P(S)$. 

To develop a connection between eq.~\eqref{eq:rearrangement_probability} and $q(S,t)$ we must take into account facilitation which will manifest itself as a particle whose softness changes without a rearrangement (i.e. because some nearby rearrangement changed its local structure). For a more detailed discussion of this derivation see the supplementary material in Schoenholz \textit{et al.}~\cite{schoenholzcubuk16}. We define the ``softness propagator'', $G(S,S_0,t)$ to be the fraction of particles that have softness $S$ at a time $t$ if they had softness $S_0$ at time $t=0$ and did not rearrange in the interval $[0,t]$. Note that $\int dSG(S,S_0,t) = 1$.

Using the softness propagator we can write down the fraction of particles that had a softness $S_0$ at $t=0$ that rearrange at a time $t$,
\begin{equation}
f_t(S) = \int dS' G(S',S,t)P_R(S').
\end{equation}
We then consider a simple model of the glass that proceeds in discreet steps of time $\tau_R$, the average duration of a rearrangement. In terms of $f_t(S)$ we can write down the softness dependent overlap function,
\begin{equation}\label{eq:softness_overlap_model}
q(S, t) = 1 - \sum_{t'=0}^t\prod_{t''=0}^{t'-1}(1-f_{t''}(S))f_{t'}(S).
\end{equation}
Thus, if we understand $P_R(S)$ and $G(S',S,t)$ then we can compute the relaxation of the entire glassy system.

To study the relaxation time we introduce a simple mean-field model where we replace our system by a system with uniform local structure with softness equal to $\langle S\rangle.$ This is equivalent to setting $P(S) = \delta(S-\langle S\rangle)$ and $G(S,S',t) = \delta (S-\langle S\rangle)$. In this approximation it follows that $q(t) = q(\langle S\rangle, t)$ and $f_t(S) = P_R(\langle S\rangle)$. It follows that,
\begin{equation}
q(t) = 1-\sum_{t'=0}^t (1-P_R(\langle S\rangle))^{t'}P_R(\langle S\rangle) = (1-P_R(\langle S\rangle))^t
\end{equation}
using the properties of geometric series. This implies that 
\begin{equation}
\tau_\alpha \sim \frac{-1}{\log(1-P_R(\langle S\rangle))}\approx \frac1{P_R(\langle S\rangle)}
\end{equation}
when $P_R(\langle S\rangle)$ is small. For our Lennard-Jones system this implies that at the level of our mean-field theory,
\begin{equation}\label{eq:relaxation_time_lj}
\tau_\alpha \sim \exp\left[-\left(\frac1{T_0} - \frac 1T\right)\left(E_0 - E_1\langle S\rangle\right)\right].
\end{equation}
If correct, this implies that the bulk of the super Arrhenius scaling of $\tau_\alpha$ comes from changes in $\langle S\rangle$.

\begin{figure}[!h]
\centering
\includegraphics[width=0.9\linewidth]{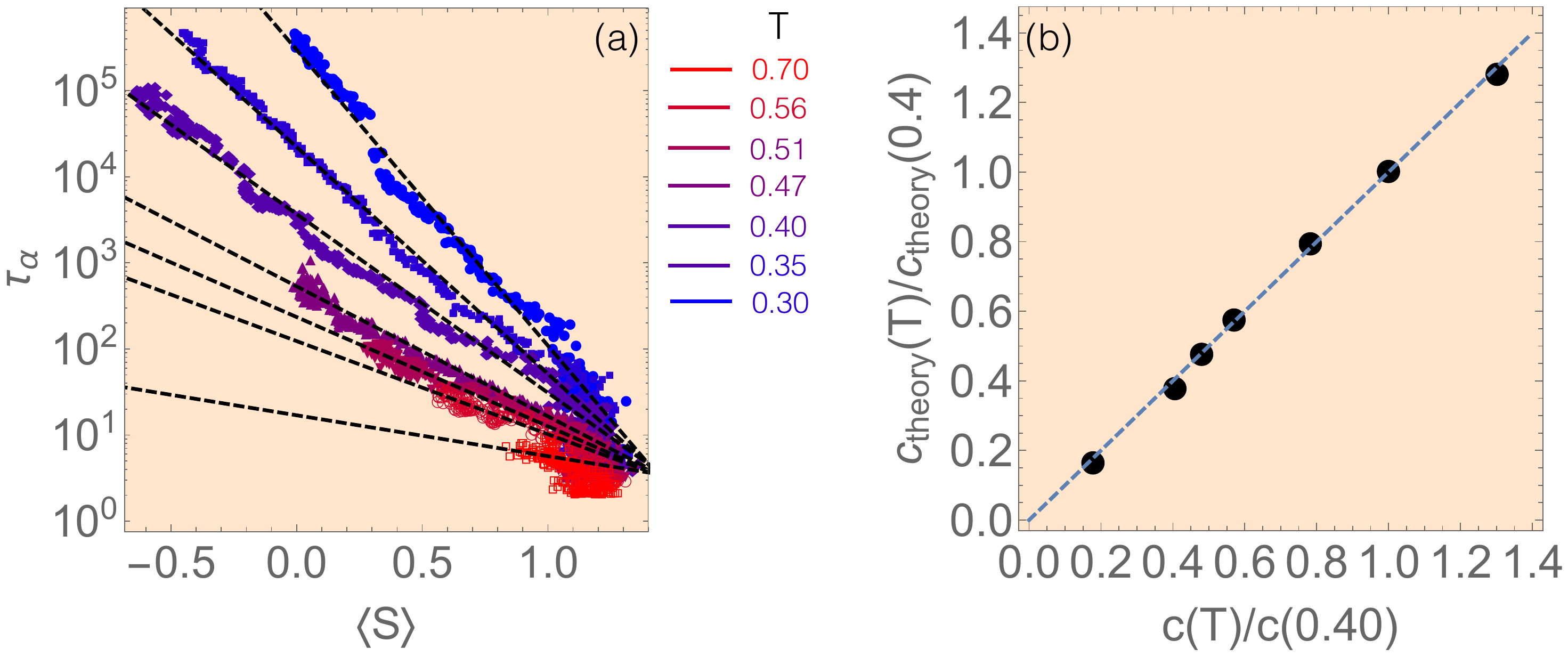}
\caption{The relaxation time. (a) The relaxation time, $\tau_\alpha$, for an aging system as a function of the average softness of the system. We see a definitive exponential relationship. (b) A comparison of the exponential pre-factor multiplying softness with the theoretical prediction. We see strong agreement between the two.}
\label{fig:relaxation}
\end{figure}

We can check the validity of this equation by considering non-equilibrium glassy systems as they age. For a more detailed discussion see Schoenholz \textit{et al.}~\cite{schoenholzcubuk17}. We let a Kob-Andersen mixture come to equilibrium at a high temperature, $T_I\approx 1.0$. and then instantaneously quench to some final temperature $T_F$. We then measure how the average softness and relaxation time evolve with time. Plotting this relationship in fig.~\ref{fig:relaxation} (a) we see that indeed the relaxation time appears to scale exponentially with the average softness, $\tau_\alpha\sim e^{-c(T)\langle S\rangle}.$  We then plot the measured ratio of $c(T_F)/c(0.4)$ against the theoretical prediction from eq.~\eqref{eq:relaxation_time_lj} for a number of different final temperatures. The result of this plot is in fig.~\ref{fig:relaxation} where we see excellent agreement.

Together these results show that our mean field model successfully captures the fundamental features of the relaxation time and moreover that eq.~\eqref{eq:relaxation_time_lj} seems to be accurate. This gives hope to mean field models of the glass transition since it seems as though the bulk relaxation time does not depend on variations in local structure. Indeed we see that, $\langle S\rangle = \int dr g(r) \tilde w(r)$ and so the relaxation time will be a function of $g(r)$ alone. Moreover, the success of eq.~\eqref{eq:relaxation_time_lj} places severe constraints on the possible forms for the functional relaxation time~\cite{schoenholzcubuk17}. It would be interesting to attempt to formally relate this formalism to mode coupling theory.

\begin{figure}[!h]
\centering
\includegraphics[width=0.9\linewidth]{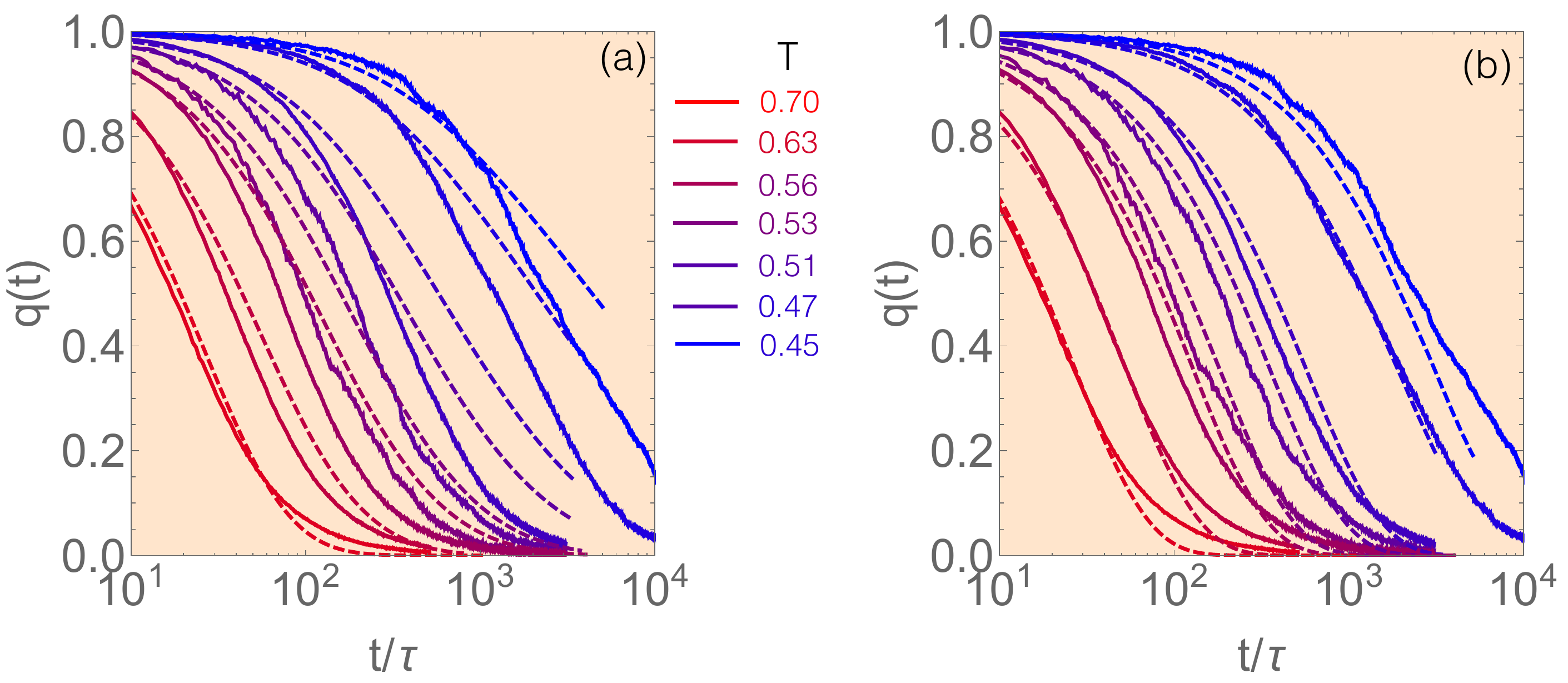}
\caption{Non-exponential behavior of the overlap function. (a) The overlap function as well as mean-field estimates from our model. The mean field relaxation clearly fails to accurately describe the behavior of $q(t)$. (b) The same plot but compared with the prediction of our full model including facilitation. There is clear agreement once local effects are taken into account.}
\label{fig:overlap}
\end{figure}

Unlike the case of the relaxation time, it appears that local structure and facilitation are necessary to properly capture the full relaxation profile of the overlap function, $q(t)$. To see this in fig.~\ref{fig:overlap} (a) we compare the overlap function found empirically to the mean field overlap function. We see that as the temperature is lowered the form of the true overlap function diverges from the mean-field prediction. In fig.~\ref{fig:overlap} (b) we plot the same overlap function against theoretical prediction found by evaluating eq.~\eqref{eq:softness_overlap_model} directly using a measured function for the softness propagator and numerically integrating it against the rearrangement probability. We see that the agreement improves drastically; indeed it seems as though the full model does successfully predict the full form of the overlap function.

\section{Conclusions and Future Directions}

We have offered a data-driven approach to understanding the behavior of supercooled liquids and glasses from their structure through the construction of the softness. We are able to use softness to connect structure to dynamics and in the process offer explanations for a number of long-standing problems in glass physics with strong experimental evidence.

The distribution of softness, the probability of rearrangement, and the softness propagator are, at this point, fundamentally empirical constructions without solid theoretical underpinning. Therefore it seems that future work should focus on making connection between these quantities and \textit{ab initio} theory. Most pressing is probably the development of a model for the dynamics of softness from more fundamental measurements. Furthermore, it would be interesting to explore the connections between the formalism presented here and the mean-field theories of random first order theory and mode-coupling theory. This is especially true of the relaxation time which we have shown explicitly does not depend on fluctuations in the local structure to an excellent approximation. Ultimately, the results presented here should act as a strong constraint on any theoretical work on the glass transition.

It is possible that at lower temperatures the Arrhenius form for $P_R(S)$ will break down. We have never been able to accurately enough estimate $P_R(S)$ at such low temperatures. An interesting direction for future work might be to probe the very low temperature behavior of softness. To do this very long simulations would have to be performed on glasses. Recent work~\cite{perez16} has shown that it might be possible to stitch together many trajectories to do very long time simulations. It would be interesting to combine this approach with softness to perform very long simulations of glasses and in turn to measure the Arrhenius form of $P_R(S)$ at very low temperatures.

Another interesting avenue for future work is to use the softness to investigate different properties of systems by measuring the distribution of energy scales and entropy scales. Already we have pursued this approach to successfully investigate the role of structure in enhancing the mobility of particles near a free surface in a polymeric glass~\cite{sussman16}. There we found minimal structural involvement which shows that these methods can be used to prove both positive and negative results about the role of structure in dynamics. We have also begun to use this method to investigate grain boundaries where it has been remarkably successful.

\section*{References}

\bibliographystyle{unsrt}
\bibliography{global_lib.bib}

\end{document}